\documentclass[letterpaper]{article} % DO NOT CHANGE THIS
\usepackage{aaai2026}  % DO NOT CHANGE THIS
\usepackage{times}  % DO NOT CHANGE THIS
\usepackage{helvet}  % DO NOT CHANGE THIS
\usepackage{courier}  % DO NOT CHANGE THIS
\usepackage[hyphens]{url}  % DO NOT CHANGE THIS
\usepackage{graphicx} % DO NOT CHANGE THIS
\usepackage{multirow}
\usepackage{makecell}
\usepackage{amsmath}
\usepackage{array}
\usepackage{booktabs}
\usepackage{natbib}  % DO NOT CHANGE THIS AND DO NOT ADD ANY OPTIONS TO IT
\usepackage{caption} % DO NOT CHANGE THIS AND DO NOT ADD ANY OPTIONS TO IT
\usepackage{algorithm}
\usepackage{algorithmic}

\usepackage{amssymb}  % mathbb
\usepackage{enumitem}

\usepackage{graphicx}

\usepackage{pifont}
\usepackage{booktabs} % For professional quality tables
\usepackage{multirow} % For multi-row cells in tables
\usepackage{graphicx} % For rotating text (optional)

\usepackage{newfloat}
\usepackage{listings}

\DeclareCaptionStyle{ruled}{labelfont=normalfont,labelsep=colon,strut=off} % DO NOT CHANGE THIS
\floatstyle{ruled}
\newfloat{listing}{tb}{lst}{}
\floatname{listing}{Listing}
\def\method{\textsc{$S^2$-KD}}

\title{$S^2$-KD: Semantic-Spectral Knowledge Distillation Spatiotemporal Forecasting}
\author {
    Wenshuo Wang\textsuperscript{\rm 1},
    Yaomin Shen\textsuperscript{\rm 2},
    Yingjie Tan\textsuperscript{\rm 3},
    Yihao Chen\textsuperscript{\rm 4}\thanks{Corresponding author.}
}
\affiliations {
    \textsuperscript{\rm 1}School of Future Technology, South China University of Technology, Guangzhou, China\\
    \textsuperscript{\rm 2}Nanchang Research Institute, Zhejiang University, Nanchang, China\\
    \textsuperscript{\rm 3}School of Software, Beihang University, Beijing, China\\
    \textsuperscript{\rm 4}College of Control Science and Engineering, Zhejiang University, Hangzhou, China\\
    202364870251@mail.scut.edu.cn, coolshennf@gmail.com, tanyingjie@buaa.edu.cn, yihaochen@zju.edu.cn
}

\usepackage{bibentry}
\begin{document}

\maketitle

\begin{abstract}
Spatiotemporal forecasting often relies on computationally intensive models to capture complex dynamics. Knowledge distillation (KD) has emerged as a key technique for creating lightweight student models, with recent advances like frequency-aware KD successfully preserving spectral properties (i.e., high-frequency details and low-frequency trends). However, these methods are fundamentally constrained by operating on pixel-level signals, leaving them blind to the rich semantic and causal context behind the visual patterns. To overcome this limitation, we introduce \textbf{S$^2$-KD}, a novel framework that unifies \textbf{S}emantic priors with \textbf{S}pectral representations for distillation. Our approach begins by training a privileged, multimodal \textbf{teacher} model. This teacher leverages textual narratives from a Large Multimodal Model (LMM) to reason about the underlying causes of events, while its architecture simultaneously decouples spectral components in its latent space. The core of our framework is a new distillation objective that transfers this unified semantic-spectral knowledge into a lightweight, \textbf{vision-only student}. Consequently, the student learns to make predictions that are not only spectrally accurate but also semantically coherent, without requiring any textual input or architectural overhead at inference. Extensive experiments on benchmarks like WeatherBench and TaxiBJ+ show that S$^2$-KD significantly boosts the performance of simple student models, enabling them to outperform state-of-the-art methods, particularly in long-horizon and complex non-stationary scenarios.
\end{abstract}

\section{Introduction}

Spatiotemporal forecasting, which aims to predict future states from historical data sequences~\cite{wu2024earthfarsser, wu2024pastnet, gao2022simvp,shi2015convolutional}, is a cornerstone of decision-making in domains ranging from climate science and meteorology to urban traffic management and autonomous navigation~\cite{wu2025triton, wang2020traffic, gao2025oneforecast}. The fundamental challenge lies in capturing the intricate coupling between high-frequency~\cite{wu2025turb, bruna2013spectral}, localized variations (e.g., sudden traffic congestion, turbulent eddies) and low-frequency~\cite{li2020fourier}, global trends (e.g., diurnal traffic patterns, seasonal climate shifts). This has spurred the development of powerful but computationally expensive models, such as complex CNN-Transformer hybrids, whose demanding resource requirements hinder their deployment in real-world, resource-constrained environments~\cite{wu2024dynst}.
\begin{figure}[!t]
	\centering
	\includegraphics[width=0.4\textwidth]{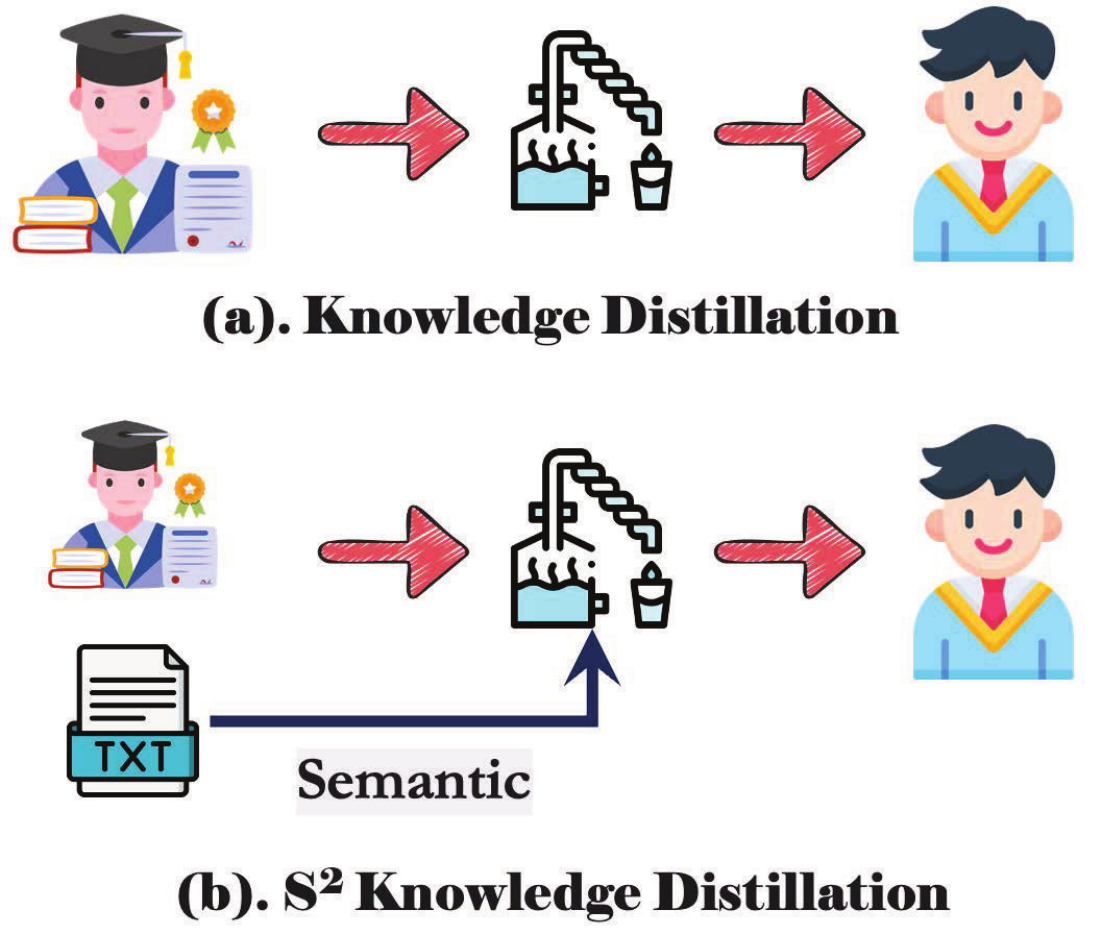}
	\caption{A comparison between traditional Knowledge Distillation and our S$^2$-KD framework. (a) Traditional KD directly distills knowledge from a teacher to a student model. (b) Our S$^2$-KD framework enriches the distillation process by injecting semantic priors extracted from text, enabling the teacher to transfer a deeper, causal understanding.}
	
	\label{fig:s2kd_concept}
\end{figure}

To address this efficiency-accuracy trade-off, knowledge distillation (KD)~\cite{gou2021knowledge, park2019relational} has emerged as a powerful paradigm for compressing large models into lightweight, efficient students \cite{chen2024effective, cicc, tcfi}. Recent advances, such as the frequency-aware distillation framework, have made significant strides by ensuring the student model preserves the spectral fidelity both high and low-frequency components of its powerful teacher. \textbf{\textit{However, these methods, despite their spectral sophistication, are fundamentally operating in a semantic vacuum.}} They are adept at mimicking \textit{what} patterns occur, but remain blind to \textit{why} they occur. For instance, they can learn the visual signature of a traffic jam, but cannot distinguish whether it is caused by predictable rush-hour volume or an unpredictable traffic accident a distinction crucial for accurate forecasting.

This reveals a critical gap in the literature: the absence of causal and semantic reasoning in current spatiotemporal distillation frameworks. \textbf{\textit{Our key insight is that natural language can serve as a powerful bridge to this missing semantic layer.}} By leveraging large multimodal models (LMM) to generate descriptive narratives of spatiotemporal scenes (e.g., \textit{ a strong cold front is approaching, causing a sudden drop in temperature}), we can provide the model with contextual and causal information at the highest level that is simply unavailable in raw pixel data. This linguistic knowledge acts as `privileged information' during training, allowing a teacher model to develop a much deeper and more robust understanding of the underlying physical processes.

To this end, we propose \textbf{\textit{S}$^2$-\textit{KD}: \textit{S}emantic-\textit{S}pectral \textit{K}nowledge \textit{D}istillation}, a novel framework designed to distill both causal understanding and spectral characteristics. S$^2$-KD first trains a powerful, multimodal \textbf{teacher} that jointly reasons over visual inputs and their corresponding textual narratives. This teacher is architected to be both \textbf{\textit{semantically-aware}} and \textbf{\textit{spectrally-decoupled}}. Subsequently, our tailored distillation process transfers this unified knowledge into a lightweight, \textbf{vision-only student}, empowering it to make semantically coherent predictions without needing any text at inference time. The student, therefore, learns to implicitly reason about causes and effects, guided by the teacher's richer, multimodal wisdom (as illustrated in Figure~\ref{fig:s2kd_concept}). Our contributions are summarized as follows:
\begin{itemize}
	\item \textbf{\textit{A New Paradigm}:} We are the first to propose a paradigm for spatiotemporal distillation that enriches representations with semantic knowledge extracted from language, moving beyond pixel-level pattern imitation to a more causal understanding.
	\item \textbf{\textit{A Novel Framework}:} We design and implement S$^2$-KD, a concrete framework featuring a multimodal teacher that fuses visual and linguistic information, and a tailored semantic-spectral distillation objective to transfer this unified knowledge to a unimodal student.
	\item \textbf{\textit{State-of-the-Art Performance}:} We conduct extensive experiments on multiple benchmarks, demonstrating that S$^2$-KD significantly boosts the performance of simple, lightweight models, enabling them to achieve new state-of-the-art results for efficient spatiotemporal forecasting.
\end{itemize}

\section{Related Work}

\noindent\textbf{\textit{Spatiotemporal Forecasting Models}} aim to capture the dynamics of complex systems. Early deep learning methods, such as ConvLSTM~\cite{shi2015convolutional} and PredRNN~\cite{wang2022predrnn}, pioneer the combination of convolution and recurrent networks to model local spatiotemporal correlations. However, they exhibit limitations in handling long-range dependencies and global dynamics~\cite{fan2020long, fahlman2022long, Sorjamaa2007MethodologyFL}. To overcome these issues, modern research shifts towards powerful CNN-Transformer hybrid architectures~\cite{chen2023contiformer, wu2024earthfarsser, chen2023fuxi, bi2023accurate, wu2025triton}. These models achieve state-of-the-art (SOTA) prediction accuracy on various benchmarks by integrating the local receptive fields of CNNs with the global modeling capabilities of Transformers. Nevertheless, this superior performance comes at the cost of immense computational and memory overhead. The self-attention mechanism in Transformers introduces quadratic complexity with respect to sequence length~\cite{wu2025triton, kurth2023fourcastnet, guibas2021adaptive}, while deep stacks of convolutions also contribute a large number of parameters. This high cost severely hinders their deployment in resource-constrained real-world scenarios, such as autonomous vehicles and edge-based weather stations, thereby highlighting the urgent need for lightweight models.

\noindent\textbf{\textit{Knowledge distillation (KD)}}~\cite{phuong2019towards, mirzadeh2020improved,stanton2021does} offers an effective pathway to address the aforementioned model complexity. Classic KD, pioneered by Hinton~\cite{hinton2015distilling}, transfers knowledge via soft labels~\cite{zhang2021graph}, while subsequent methods like FitNets focus on matching intermediate features~\cite{murata2023recurrent}. In the spatiotemporal domain, KD has also seen significant progress. In particular, frequency-aware distillation frameworks like SDKD represent a major advancement by preserving the spectral fidelity of the teacher model, including both high and low-frequency components. \textit{However, a fundamental limitation underlies all existing KD approaches, from classic to frequency-aware: they operate exclusively within a unimodal paradigm.} This means they can only distill the visual patterns the teacher model `sees' (\textit{what}) but fail to transfer the underlying causal understanding of these patterns (\textit{why}). They distill `appearance' rather than `comprehension', making the student model vulnerable to dynamic changes caused by unseen factors (e.g., sudden events), even if it is spectrally aligned with the teacher.

\noindent\textbf{\textit{Multimodal Learning and Privileged Information.}} To break the semantic bottleneck of unimodal distillation, our work draws inspiration from multimodal learning and the concept of privileged information~\cite{ramachandram2017deep, blikstein2013multimodal}. Recent breakthroughs in Large Multimodal Models (LMMs), such as CLIP~\cite{radford2021learning} and LLaVA~\cite{liu2023visual}, successfully bridge the gap between vision and language, enabling the generation of high-level, logically coherent textual descriptions for spatiotemporal scenes. Natural language not only provides a holistic summary of a scene but, more importantly, contains rich causal relationships, object attributes, and common-sense knowledge that is difficult to extract from raw pixel data alone. We frame our approach within Vapnik's paradigm of Learning Using Privileged Information (LUPI)~\cite{pechyony2010theory}. In this framework, the textual narratives generated by LMMs serve as the `privileged information', which is available only during the training phase to help the teacher model build a profound understanding of the dynamics. The S$^2$-KD framework, therefore, essentially distills this deep understanding gained from privileged information and internalizes it within the parameters of a vision-only student model through a novel semantic-spectral distillation process. Unlike previous works that focus exclusively on knowledge transfer within a single modality, our work is the first to explore how to distill cross-modal semantic knowledge from a multimodal privileged teacher to a unimodal student, opening up a new avenue for building more intelligent and robust lightweight forecasting models.

\section{Methodology}
\begin{figure*}[!t]
	\centering
	\includegraphics[width=1\textwidth]{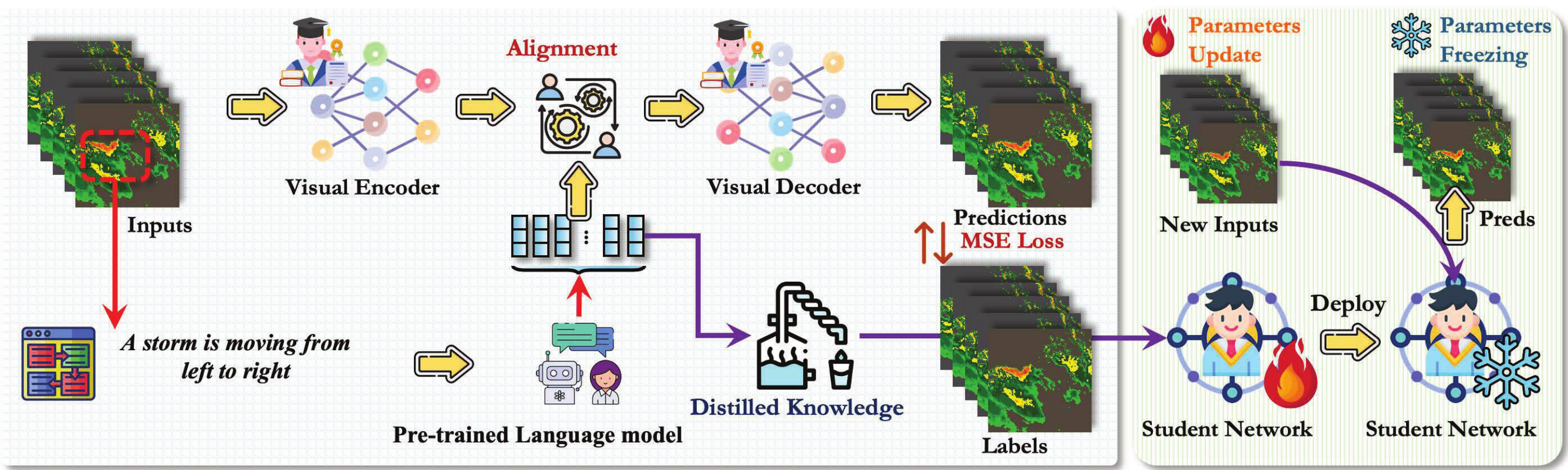} 
	\caption{An overview of our proposed S$^2$-KD framework.}
		\label{fig:s2kd_framework}
	\end{figure*}
	
	\subsection{Problem Formulation}
	
	The primary goal of spatiotemporal forecasting is to predict a sequence of future states given a history of observations. Let $\mathcal{X} = \{\mathbf{X}_t \in \mathbb{R}^{H \times W \times C}\}_{t=1}^{T_{in}}$ represent the historical sequence of spatiotemporal data, where $H$ and $W$ are the spatial dimensions, $C$ is the number of channels, and $T_{in}$ is the length of the input sequence. The objective is to learn a mapping function $\mathcal{F}$ that predicts the future sequence $\mathcal{Y} = \{\mathbf{Y}_{t'} \in \mathbb{R}^{H \times W \times C}\}_{t'=1}^{T_{out}}$, where $T_{out}$ is the prediction horizon. A conventional forecasting model is trained by minimizing an objective function, typically the Mean Squared Error (MSE), between the predictions and the ground truth:
	\begin{equation}
		\mathcal{L}_{pred} = \mathbb{E}_{(\mathcal{X}, \mathcal{Y})} \left\| \mathcal{F}(\mathcal{X}; \theta_F) - \mathcal{Y} \right\|_2^2,
	\end{equation}
	where $\theta_F$ are the parameters of the model $\mathcal{F}$. Our work extends this formulation by introducing a knowledge distillation framework. We first train a powerful multimodal teacher model $\mathcal{T}$, which leverages both the visual sequence $\mathcal{X}$ and a corresponding textual description $\mathcal{S}$ generated by a Large Multimodal Model (LMM). This teacher is optimized to produce highly accurate predictions. Subsequently, we aim to train a lightweight, vision-only student model $\mathcal{G}(\mathcal{X}; \theta_G)$ to mimic the teacher's behavior. The student's training objective is a composite loss that includes not only the prediction loss but also a distillation loss, which transfers the semantic-spectral knowledge from the teacher:
	\begin{equation}
		\min_{\theta_G} \mathcal{L}_{pred}(\mathcal{G}(\mathcal{X}), \mathcal{Y}) + \lambda \mathcal{L}_{distill}(\mathcal{G}(\mathcal{X}), \mathcal{T}(\mathcal{X}, \mathcal{S})),
		\label{eq:overall_objective}
	\end{equation}
	where $\mathcal{L}_{distill}$ measures the discrepancy between the student's and teacher's internal representations, and $\lambda$ is a hyperparameter balancing the two loss terms. The ultimate goal is to obtain an efficient student model $\mathcal{G}$ that achieves performance comparable to the privileged teacher $\mathcal{T}$ at inference time without requiring the textual input $\mathcal{S}$.
	
	\subsection{Overall Architecture of S$^2$-KD}
	
	Our S$^2$-KD framework, illustrated in Figure~\ref{fig:s2kd_framework}, comprises two stages: training a privileged multimodal teacher for knowledge distillation, and deploying a lightweight student for efficient inference. During training, the teacher model $\mathcal{T}$ processes both the visual sequence $\mathcal{X}$ and a textual narrative $\mathcal{S}$ from a Large Multimodal Model (LMM). An internal \textit{Alignment Module} fuses these inputs into a unified, semantically-rich representation, guided by the ground truth $\mathcal{Y}$. Instead of distilling from final predictions, we extract knowledge directly from this fused intermediate representation. We define this as \textit{semantic-spectral knowledge}, as it combines causal semantics from language with the spectral properties of the visual dynamics. Subsequently, a lightweight student model $\mathcal{G}$, which only sees $\mathcal{X}$, is trained to mimic this knowledge via the objective in Equation~\ref{eq:overall_objective}. This distillation process internalizes the teacher's multimodal reasoning into the student's parameters. As a result, the deployed student performs efficient inference on visual data alone, yet retains the teacher's semantic understanding. \textbf{\textit{The design of this framework ensures that the complexity of multimodal reasoning is confined to the offline training stage, guaranteeing high efficiency for online deployment}.}
	
	\subsection{Multimodal Privileged Teacher}
	
	Our privileged teacher model, $\mathcal{T}$, is architected to effectively fuse spatiotemporal visual patterns with high-level semantic narratives. It comprises a visual encoder $\mathcal{E}_v$, a text encoder $\mathcal{E}_s$, a cross-modal alignment module, and a predictive decoder $\mathcal{D}_v$. The visual encoder, $\mathcal{E}_v$, first maps the input visual sequence $\mathcal{X}$ into a sequence of latent feature vectors:
	\begin{equation}
		\mathbf{Z}_v = \mathcal{E}_v(\mathcal{X}), \quad \text{where } \mathbf{Z}_v \in \mathbb{R}^{L_v \times D}.
	\end{equation}
	Simultaneously, the text encoder $\mathcal{E}_s$, a pre-trained Transformer, processes the textual description $\mathcal{S}$ to produce a sequence of semantic embeddings:
	\begin{equation}
		\mathbf{Z}_s = \mathcal{E}_s(\mathcal{S}), \quad \text{where } \mathbf{Z}_s \in \mathbb{R}^{L_s \times D}.
	\end{equation}
	The core of our teacher is the \textit{Cross-Modal Alignment Module}, which facilitates deep interaction between these two modalities using a stack of $N$ cross-attention layers. For each layer $i \in \{1, ..., N\}$, we first compute the query ($\mathbf{Q}$), key ($\mathbf{K}$), and value ($\mathbf{V}$) matrices from the visual and text features of the previous layer, $\mathbf{Z}_v^{(i-1)}$ and $\mathbf{Z}_s^{(i-1)}$ (with $\mathbf{Z}_v^{(0)} = \mathbf{Z}_v, \mathbf{Z}_s^{(0)} = \mathbf{Z}_s$), using learnable projection matrices:
	\begin{equation}
		\mathbf{Q}^{(i)} = \mathbf{Z}_v^{(i-1)} \mathbf{W}_Q^{(i)},
	\end{equation}
	\begin{equation}
		\mathbf{K}^{(i)} = \mathbf{Z}_s^{(i-1)} \mathbf{W}_K^{(i)},
	\end{equation}
	\begin{equation}
		\mathbf{V}^{(i)} = \mathbf{Z}_s^{(i-1)} \mathbf{W}_V^{(i)},
	\end{equation}
	where $\mathbf{W}_Q^{(i)}, \mathbf{W}_K^{(i)}, \mathbf{W}_V^{(i)} \in \mathbb{R}^{D \times D}$ are the projection parameters. The semantically-enhanced features are then computed via multi-head attention (MHA):
	\begin{equation}
		\text{Attn}(\mathbf{Q}^{(i)}, \mathbf{K}^{(i)}, \mathbf{V}^{(i)})=\text{Softmax}\left(\frac{\mathbf{Q}^{(i)}{\mathbf{K}^{(i)}}^T}{\sqrt{d_k}}\right)\mathbf{V}^{(i)}.
	\end{equation}
	This operation is followed by a residual connection and layer normalization, forming a complete cross-attention block. The output of the $i$-th block, $\mathbf{Z}_v^{(i)}$, is then fed into the next. After $N$ such blocks, we obtain the final fused representation $\mathbf{Z}_{\text{fused}} = \mathbf{Z}_v^{(N)}$, which serves as the source for distillation:
	\begin{equation}
		\mathbf{Z}_{\text{fused}} = \text{LayerNorm}\left(\mathbf{Z}_v^{(i-1)} + \text{MHA}(\mathbf{Q}^{(i)}, \mathbf{K}^{(i)}, \mathbf{V}^{(i)})\right).
		\label{eq:fused_representation_detailed}
	\end{equation}
	Finally, the predictive decoder $\mathcal{D}_v$ takes this fused representation $\mathbf{Z}_{\text{fused}}$ and reconstructs the future spatiotemporal sequence $\hat{\mathcal{Y}} = \mathcal{D}_v(\mathbf{Z}_{\text{fused}})$.
	
	\subsection{Semantic-Spectral Distillation Loss}
	
	Having defined the powerful multimodal teacher, the next crucial step is to design a distillation loss, $\mathcal{L}_{distill}$, that effectively transfers its rich, unified knowledge to the lightweight student $\mathcal{G}$. Our objective is to formulate a loss that preserves both the high-level semantic understanding and the fine-grained spectral characteristics captured in the teacher's fused representation, $\mathbf{Z}_{\text{fused}}^{\mathcal{T}}$, from Equation~\ref{eq:fused_representation_detailed}. To achieve this, we design a composite loss consisting of two complementary components: a semantic alignment loss and a spectral alignment loss.
	
	First, we align the intermediate representations of the student and teacher. The student model $\mathcal{G}$ employs a visual encoder $\mathcal{E}_v^{\mathcal{G}}$ to generate its own latent representation $\mathbf{Z}_{v}^{\mathcal{G}} = \mathcal{E}_v^{\mathcal{G}}(\mathcal{X})$. To ensure dimensional compatibility with the teacher's representation, we apply a linear projection layer $P$. The core of our distillation is to minimize the discrepancy between the student's projected representation, $P(\mathbf{Z}_{v}^{\mathcal{G}})$, and the teacher's fused representation, $\mathbf{Z}_{\text{fused}}^{\mathcal{T}}$.
	
	The first component is the \textbf{semantic alignment loss}, $\mathcal{L}_{semantic}$, which enforces the student to capture the high-level semantic structure of the teacher's representation. We employ the Mean Squared Error (MSE) for this purpose, as it effectively matches the overall feature distributions:
	\begin{equation}
		\mathcal{L}_{semantic} = \left\| P(\mathbf{Z}_{v}^{\mathcal{G}}) - \mathbf{Z}_{\text{fused}}^{\mathcal{T}} \right\|_2^2.
	\end{equation}
	This loss compels the student to reconstruct the teacher's semantically-informed "thought process," thereby implicitly internalizing the knowledge gained from language.
	
	The second component is the \textbf{\textit{Spectral Alignment Loss}}, $\mathcal{L}_{spectral}$, which inherits the core idea from frequency-aware distillation to preserve the modeling of both high-frequency details and low-frequency trends. We apply the Fast Fourier Transform ($\mathcal{F}$) to map the features into the frequency domain and compute the L1 loss between their spectral magnitudes:
	\begin{equation}
		\mathcal{L}_{spectral} = \left\| \left|\mathcal{F}\left(P(\mathbf{Z}_{v}^{\mathcal{G}})\right)\right| - \left|\mathcal{F}\left(\mathbf{Z}_{\text{fused}}^{\mathcal{T}}\right)\right| \right\|_1.
	\end{equation}
	By directly aligning the spectral representations, we explicitly guide the student to learn the same frequency response as the teacher. Finally, our total distillation loss is a weighted sum of these two components:
	\begin{equation}
		\mathcal{L}_{distill} = \mathcal{L}_{semantic} + \beta \mathcal{L}_{spectral},
	\end{equation}
	where $\beta$ is a hyperparameter balancing the importance of semantic and spectral alignment. This composite loss ensures a comprehensive knowledge transfer, encompassing both macroscopic semantic understanding and microscopic dynamic details.
	
	\subsection{Final Objective and Training Procedure}
	
	The training of our S$^2$-KD framework is conducted in two sequential stages: first, pre-training the multimodal privileged teacher, and second, training the lightweight student via our proposed semantic-spectral distillation.
	
	\paragraph{Stage 1: Teacher Model Training.}
	In the first stage, we train the multimodal teacher model $\mathcal{T}$ to learn an effective mapping from the visual sequence $\mathcal{X}$ and the textual narrative $\mathcal{S}$ to the future sequence $\mathcal{Y}$. The teacher is optimized solely based on the standard predictive loss, which is the Mean Squared Error (MSE) between its predictions and the ground truth:
	\begin{equation}
		\mathcal{L}_{\text{Teacher}} = \mathbb{E}_{(\mathcal{X}, \mathcal{S}, \mathcal{Y})} \left\| \mathcal{T}(\mathcal{X}, \mathcal{S}; \theta_T) - \mathcal{Y} \right\|_2^2.
	\end{equation}
	Through this process, the teacher learns to leverage semantic information to form a high-quality internal representation $\mathbf{Z}_{\text{fused}}^{\mathcal{T}}$. Upon completion of this stage, the parameters $\theta_T$ of the teacher model are frozen.
	
	\paragraph{Stage 2: Student Model Distillation.}
	In the second stage, we train the lightweight, vision-only student model $\mathcal{G}$. The student is optimized using a composite objective function that combines the predictive loss with our semantic-spectral distillation loss $\mathcal{L}_{distill}$. The final objective for the student model is:
	\begin{equation}
		\min_{\theta_G} \mathcal{L}_{\text{Student}} = \mathcal{L}_{pred}(\mathcal{G}(\mathcal{X}), \mathcal{Y}) + \lambda \mathcal{L}_{distill},
	\end{equation}
	where $\mathcal{L}_{pred}$ is the standard MSE loss for the student's predictions, and $\mathcal{L}_{distill}$ is defined as $\mathcal{L}_{semantic} + \beta \mathcal{L}_{spectral}$. The hyperparameters $\lambda$ and $\beta$ balance the contributions of the predictive task, semantic alignment, and spectral alignment. During this stage, the teacher model operates in evaluation mode solely to provide the target representation $\mathbf{Z}_{\text{fused}}^{\mathcal{T}}$, and no gradients are backpropagated through it. Upon completion of this two-stage procedure, we obtain a lightweight and efficient student model $\mathcal{G}$ that inherits the advanced reasoning capabilities of the privileged teacher, ready for deployment.
	
\section{Experiment}
We aim to answer three key research questions:
\begin{itemize}[leftmargin=*]
	\item \textbf{RQ1.} How does \method{} perform against state-of-the-art forecasting and knowledge distillation methods?
	\item \textbf{RQ2.} What are the individual contributions of the semantic and spectral distillation components?
	\item \textbf{RQ3.} How effective is \method{} in predicting high-impact extreme events?
\end{itemize}

\subsection{Experimental Setup}

\paragraph{Datasets.}
We evaluate \method{} on three benchmarks to assess its performance and generalization capabilities. \textbf{\ding{182}. Prometheus} is a large-scale fire simulation dataset designed for out-of-distribution (OOD) fluid dynamics. It includes two scenarios, Tunnel Fire (Prometheus-T)~\cite{wu2024prometheus} and Pool Fire (Prometheus-P), with distribution shifts created by varying physical parameters like heat release rate. Models are trained on seen environments and tested on unseen ones to evaluate OOD generalization. \textbf{\ding{183}. WeatherBench (ERA5)}~\cite{rasp2020weatherbench} is a scientific benchmark for global weather forecasting. We use Geopotential (Z500) and Temperature (T850) variables to capture large-scale, slowly-varying atmospheric dynamics. \textbf{\ding{184}. TaxiBJ+}~\cite{wu2023earthfarseer} is an urban traffic flow dataset from Beijing. It represents a highly non-stationary system, challenging models to capture both periodic patterns and stochastic events.

\paragraph{Model Selection.}
To demonstrate the versatility and effectiveness of our \method{} framework, we adopt a domain-specific teacher and general-purpose student strategy. 
For each benchmark, we select a powerful, state-of-the-art model from its respective domain as the \textbf{teacher}: \textbf{Triton}~\cite{wu2025triton} for weather forecasting on WeatherBench, \textbf{EarthFarseer}~\cite{wu2023earthfarseer} for urban dynamics on TaxiBJ+, and a deep variant of \textbf{SimVP}~\cite{gao2022simvp} for fluid dynamics on Prometheus. This ensures that the distilled knowledge originates from a top-performing specialist. 
For semantic extraction, we primarily use the open-source \textbf{DeepSeek-VL}~\cite{lu2024deepseek}, with other LLMs explored in our ablation studies. 
For the \textbf{student}, we employ a diverse set of lightweight architectures, including \textbf{U-Net}~\cite{ronneberger2015u}, a shallow \textbf{ResNet}~\cite{he2016deep}, and an \textbf{MLP-Mixer}~\cite{tolstikhin2021mlp}, to validate that the benefits of \method{} are architecture-agnostic. 
As for distillation baselines, we compare against the classic \textbf{Standard KD}~\cite{hinton2015distilling} and feature-based \textbf{FitNet}~\cite{chen2021distilling}.

\paragraph{Implementation Details.}
All experiments are conducted on a server equipped with four NVIDIA A100 (80GB) GPUs, using PyTorch 2.1 and CUDA 12.0. We employ the Adam optimizer~\cite{kingma2014adam} for all model training. For the teacher models, we largely follow the optimal hyperparameter configurations reported in their original papers. For student model training via distillation, we set an initial learning rate of $1 \times 10^{-3}$, which is reduced by a factor of 10 if the validation loss plateaus for 5 consecutive epochs. The batch size is set to 16. All models are trained for a maximum of 100 epochs with an early stopping mechanism based on the validation set performance to prevent overfitting. For our \method{} framework, the distillation loss weight $\lambda$ is set to 1.0, and the spectral alignment weight $\beta$ is set to 0.5, determined via a grid search on a validation subset. For the LMM-based semantic extraction, we generate a single descriptive caption for each input sequence using a standardized prompt template. Unless otherwise specified, DeepSeek-VL is used as the default LLM. To ensure reproducibility, we set the global random seed to 42 for all experiments.
\begin{table}[ht]
	\centering
	\resizebox{\columnwidth}{!}{%
		\begin{tabular}{@{}llccccc@{}}
			\toprule
			\textbf{Model} & \textbf{Method} & \textbf{Params} & \textbf{Latency} & \textbf{MSE} & \textbf{MAE} & \textbf{SSIM} \\
			& & \textbf{(M)} $\downarrow$ & \textbf{(ms)} $\downarrow$ & $\downarrow$ & $\downarrow$ & $\uparrow$ \\ \midrule
			
			\textbf{Teacher (Triton)} & - & \textbf{150.2} & \textbf{85.6} & \textbf{0.0683} & \textbf{0.7287} & \textbf{0.9493} \\ \midrule
			\multirow{3}{*}{\textbf{U-Net}} & Baseline & 5.1 & 12.3 & 0.0831 & 0.9822 & 0.8635 \\
			& + FitNet & 5.1 & 12.3 & 0.0765 & 0.9150 & 0.8712 \\
			& \textbf{+ \method{} (Ours)} & \textbf{5.1} & \textbf{12.3} & \textbf{0.0698} & \textbf{0.8104} & \textbf{0.9012} \\ \midrule
			\multirow{3}{*}{\textbf{ResNet}} & Baseline & 4.5 & 10.8 & 0.0876 & 1.0210 & 0.8590 \\
			& + FitNet & 4.5 & 10.8 & 0.0801 & 0.9433 & 0.8705 \\
			& \textbf{+ \method{} (Ours)} & \textbf{4.5} & \textbf{10.8} & \textbf{0.0753} & \textbf{0.8819} & \textbf{0.8831} \\ \midrule
			\multirow{3}{*}{\textbf{MLP-Mixer}} & Baseline & 6.2 & 15.1 & 0.0953 & 1.1527 & 0.8421 \\
			& + FitNet & 6.2 & 15.1 & 0.0925 & 1.1098 & 0.8490 \\
			& \textbf{+ \method{} (Ours)} & \textbf{6.2} & \textbf{15.1} & \textbf{0.0895} & \textbf{1.0436} & \textbf{0.8615} \\ 
			\bottomrule
		\end{tabular}%
	}
	\caption{Performance on \textbf{WeatherBench}. Teacher model (Triton) results are provided as an upper bound. Our \method{} consistently enables lightweight students to approach the teacher's performance more closely than other methods, including the classic feature-based baseline (FitNet).}
	\label{tab:weatherbench_main_results}
\end{table}

\begin{table}[ht]
	\centering
	\resizebox{\columnwidth}{!}{%
		\begin{tabular}{@{}llccccc@{}}
			\toprule
			\textbf{Model} & \textbf{Method} & \textbf{Params} & \textbf{Latency} & \textbf{MSE} & \textbf{MAE} & \textbf{SSIM} \\
			& & \textbf{(M)} $\downarrow$ & \textbf{(ms)} $\downarrow$ & $\downarrow$ & $\downarrow$ & $\uparrow$ \\ \midrule
			
			\textbf{Teacher (EarthFarseer)} & - & \textbf{125.8} & \textbf{72.4} & \textbf{0.1172} & \textbf{0.9701} & \textbf{0.9810} \\ \midrule
			\multirow{3}{*}{\textbf{U-Net}} & Baseline & 5.1 & 12.3 & 0.1354 & 1.1032 & 0.9532 \\
			& + FitNet & 5.1 & 12.3 & 0.1298 & 1.0760 & 0.9580 \\
			& \textbf{+ \method{} (Ours)} & \textbf{5.1} & \textbf{12.3} & \textbf{0.1180} & \textbf{0.9855} & \textbf{0.9754} \\ \midrule
			\multirow{3}{*}{\textbf{ResNet}} & Baseline & 4.5 & 10.8 & 0.1402 & 1.1450 & 0.9499 \\
			& + FitNet & 4.5 & 10.8 & 0.1331 & 1.1027 & 0.9556 \\
			& \textbf{+ \method{} (Ours)} & \textbf{4.5} & \textbf{10.8} & \textbf{0.1231} & \textbf{1.0189} & \textbf{0.9678} \\ \midrule
			\multirow{3}{*}{\textbf{MLP-Mixer}} & Baseline & 6.2 & 15.1 & 0.1520 & 1.2345 & 0.9380 \\
			& + FitNet & 6.2 & 15.1 & 0.1485 & 1.2010 & 0.9413 \\
			& \textbf{+ \method{} (Ours)} & \textbf{6.2} & \textbf{15.1} & \textbf{0.1399} & \textbf{1.1401} & \textbf{0.9523} \\
			\bottomrule
		\end{tabular}%
	}
	\caption{Performance on \textbf{TaxiBJ+}. For this non-stationary urban dynamics task, our \method{} not only surpasses other methods but also brings the lightweight students remarkably close to the performance of the large EarthFarseer teacher model.}
	\label{tab:taxibj_main_results}
\end{table}

\begin{table}[ht]
	\centering
	\resizebox{\columnwidth}{!}{%
		\begin{tabular}{@{}llccccc@{}}
			\toprule
			\textbf{Model} & \textbf{Method} & \textbf{Params} & \textbf{Latency} & \textbf{MSE} & \textbf{MAE} & \textbf{SSIM} \\
			& & \textbf{(M)} $\downarrow$ & \textbf{(ms)} $\downarrow$ & $\downarrow$ & $\downarrow$ & $\uparrow$ \\ \midrule
			
			\textbf{Teacher (SimVP-Deep)} & - & \textbf{80.5} & \textbf{55.1} & \textbf{0.0210} & \textbf{0.1805} & \textbf{0.8521} \\ \midrule
			\multirow{3}{*}{U-Net} & Baseline & 5.1 & 12.3 & 0.0295 & 0.2510 & 0.7811 \\
			& + FitNet & 5.1 & 12.3 & 0.0268 & 0.2243 & 0.8034 \\
			& \textbf{+ \method{} (Ours)} & \textbf{5.1} & \textbf{12.3} & \textbf{0.0219} & \textbf{0.1895} & \textbf{0.8416} \\ \midrule
			\multirow{3}{*}{ResNet} & Baseline & 4.5 & 10.8 & 0.0312 & 0.2680 & 0.7725 \\
			& + FitNet & 4.5 & 10.8 & 0.0280 & 0.2415 & 0.7992 \\
			& \textbf{+ \method{} (Ours)} & \textbf{4.5} & \textbf{10.8} & \textbf{0.0240} & \textbf{0.2033} & \textbf{0.8291} \\ \midrule
			\multirow{3}{*}{MLP-Mixer} & Baseline & 6.2 & 15.1 & 0.0350 & 0.2915 & 0.7504 \\
			& + FitNet & 6.2 & 15.1 & 0.0321 & 0.2706 & 0.7810 \\
			& \textbf{+ \method{} (Ours)} & \textbf{6.2} & \textbf{15.1} & \textbf{0.0286} & \textbf{0.2407} & \textbf{0.8077} \\
			\bottomrule
		\end{tabular}%
	}
	\caption{Performance on the \textbf{Prometheus} (OOD) dataset. The results show that our \method{} provides the best generalization to unseen physical conditions, achieving performance closest to the teacher model with a fraction of the computational cost.}
	\label{tab:prometheus_main_results}
\end{table}

\subsection{Main results (RQ1.)}

This section addresses our first research question (RQ1): \textit{How does \method{} perform compared to the baseline and classic knowledge distillation methods?} We evaluate our framework through extensive experiments on three benchmark datasets with diverse dynamics, WeatherBench, TaxiBJ + and Prometheus, with the performance of teacher models serving as a reference upper bound.

The detailed results in Tables~\ref{tab:weatherbench_main_results}, \ref{tab:taxibj_main_results}, and \ref{tab:prometheus_main_results} clearly demonstrate the effectiveness of our approach. While classic knowledge distillation methods like FitNet consistently improve upon the baseline models, our \method{} framework achieves a far more substantial leap in performance. This superiority is evident across all datasets; for instance, the \method{} empowered U-Net shows a 9.0\% MSE improvement over FitNet on WeatherBench. This suggests that for complex spatiotemporal dynamics, simple feature mimicry is suboptimal, whereas the structured semantic and spectral knowledge provided by \method{} offers a more potent guidance signal. Moreover, this performance advantage is not confined to a single architecture. The superiority of our framework holds consistently across diverse student models, including the convolution-based U-Net and ResNet, and the MLP-based Mixer, robustly demonstrating that \method{} is a general, architecture-agnostic framework. Ultimately, \method{} enables lightweight student models with only a few million parameters to achieve performance remarkably close to their massive teacher counterparts. On the TaxiBJ+ dataset, the distilled U-Net (MSE of 0.1180) nearly matches the performance of the 125.8M-parameter teacher (MSE of 0.1172), highlighting the framework's exceptional capability in balancing high performance with computational efficiency. In summary, the experimental results provide compelling evidence for the superiority of the \method{} framework, establishing it as a powerful and versatile solution that not only surpasses traditional distillation methods but also consistently elevates the performance of lightweight models to new heights across various tasks and architectures.

\subsection{Ablation Study (RQ2.)}

To answer our second research question (RQ2): \textit{What are the individual contributions of the semantic and spectral distillation components?} we conduct a series of ablation experiments to quantitatively dissect the effectiveness of each core component within our \method{} framework. This study is performed with the U-Net as the student model across three datasets with distinct characteristics.

\begin{figure}[t]
	\centering
	\includegraphics[width=\columnwidth]{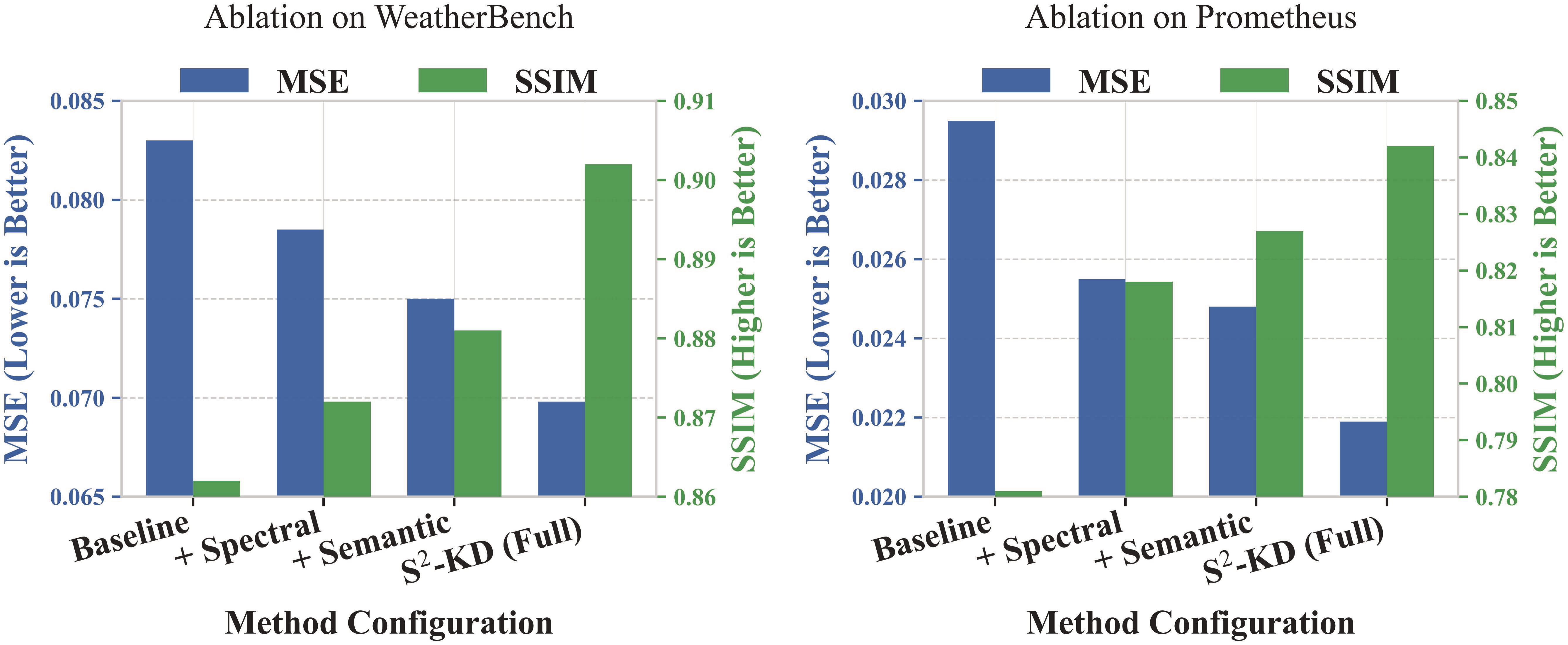} 
	\caption{Ablation study of \method{} on (a) WeatherBench and (b) Prometheus. Both MSE (blue, left axis) and SSIM (green, right axis) metrics demonstrate that while each component is individually beneficial, their combination in the full \method{} framework yields the best performance. This validates the synergistic effect of integrating semantic and spectral knowledge.}
	\label{fig:ablation}
\end{figure}

The experimental results are clearly presented in Table~\ref{tab:ablation_study} and Figure~\ref{fig:ablation}, from which a highly consistent trend can be observed.

First, introducing either the spectral distillation loss (+ Spectral) or the semantic distillation loss (+ Semantic) alone improves performance over the baseline model across all metrics. For instance, on the TaxiBJ+ dataset (Table~\ref{tab:ablation_study}), semantic distillation alone reduces the MSE from 0.1354 to 0.1261. This provides strong evidence for the individual effectiveness of our framework's two core components.

Second, the full \method{} framework, which integrates both components, achieves the best performance in all tested scenarios. As illustrated in Figure~\ref{fig:ablation}, on both WeatherBench and Prometheus, the full method not only secures the lowest MSE but also attains the highest SSIM. This result compellingly demonstrates a significant synergistic effect between the semantic and spectral knowledge. They are not merely additive but complementary, and their combination is essential for maximizing the student model's performance, thus validating the rationale behind our method's design.

\begin{table}[ht]
	\centering
	\resizebox{\columnwidth}{!}{%
		\begin{tabular}{@{}cccc|ccc@{}}
			\toprule
			\multicolumn{4}{c|}{\textbf{Method Components}} & \multicolumn{3}{c}{\textbf{Performance Metrics}} \\ \midrule
			\textbf{Baseline} & \textbf{+ $\mathcal{L}_{pred}$} & \textbf{+ $\mathcal{L}_{spectral}$} & \textbf{+ $\mathcal{L}_{semantic}$} & \textbf{MSE} $\downarrow$ & \textbf{MAE} $\downarrow$ & \textbf{SSIM} $\uparrow$ \\ \midrule
			\checkmark & \checkmark & & & 0.1354 & 1.1032 & 0.9532 \\
			\checkmark & \checkmark & \checkmark & & 0.1280 & 1.0617 & 0.9591 \\
			\checkmark & \checkmark & & \checkmark & 0.1261 & 1.0455 & 0.9610 \\
			
			\checkmark & \checkmark & \checkmark & \checkmark & \textbf{0.1180} & \textbf{0.9855} & \textbf{0.9754} \\ \bottomrule
		\end{tabular}%
	}
	\caption{Ablation study of \method{} components on the TaxiBJ+ dataset with a U-Net student.  The results demonstrate that both semantic and spectral distillation are beneficial, and their combination yields a synergistic effect, leading to the best overall performance.}
	\label{tab:ablation_study}
\end{table}

\subsection{Analysis on Different Large Language Models}
A core hypothesis of our \method{} framework is that high-quality textual narratives provide invaluable semantic and causal knowledge to the teacher model. To validate this and investigate the framework's sensitivity to the quality of the language model, we conduct a comparative experiment. We select three representative Large Multimodal Models (LMMs) to generate textual descriptions and compare their results against a baseline (FitNet) that uses no semantic priors.

\begin{figure}[t]
	\centering
	\includegraphics[width=\columnwidth]{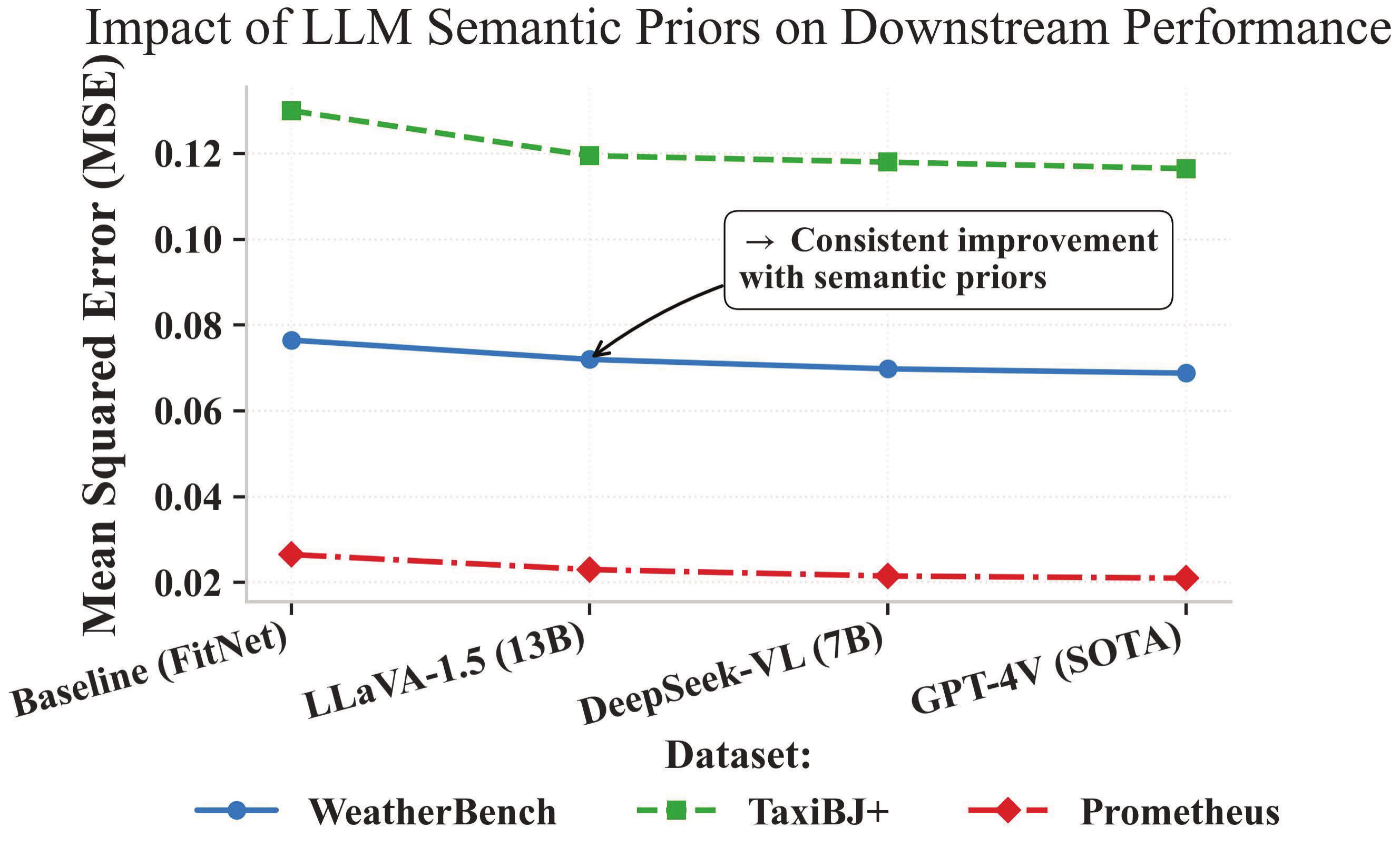} 
	\caption{Impact of different LMMs on the performance of the U-Net student across three datasets. The results show a positive correlation between LMM capability and student performance (MSE). Meanwhile, the largest performance leap occurs when moving from no semantic prior (Baseline) to using any LMM, demonstrating the robustness and practicality of our framework.}
	\label{fig:llm_semantic_impact}
	
\end{figure}

The results, illustrated in Figure~\ref{fig:llm_semantic_impact}, reveal two important conclusions. First, the final performance of the student model positively correlates with the capability of the LMM providing the text. Across all three datasets, we observe a clear downward trend: as the LMM progresses from the open-source LLaVA-1.5 and DeepSeek-VL to the state-of-the-art GPT-4V, the student model's MSE consistently decreases. This provides strong evidence that higher-quality, more insightful textual descriptions indeed translate into more effective knowledge, thereby enhancing student performance.

Second, and equally important, the most significant performance gain occurs in the leap from having no semantic prior (Baseline) to having any semantic prior (LLaVA-1.5). This indicates that our framework is robust and not fragilely dependent on a single, top-tier model. Even with moderately-sized, open-source models, \method{} delivers substantial benefits far exceeding traditional methods. This analysis not only confirms the importance of semantic knowledge quality but also demonstrates the practicality and adaptability of the \method{} framework as a general approach.

\subsection{Qualitative Analysis (RQ3.)}

\begin{figure}[t]
	\centering
	\includegraphics[width=\columnwidth]{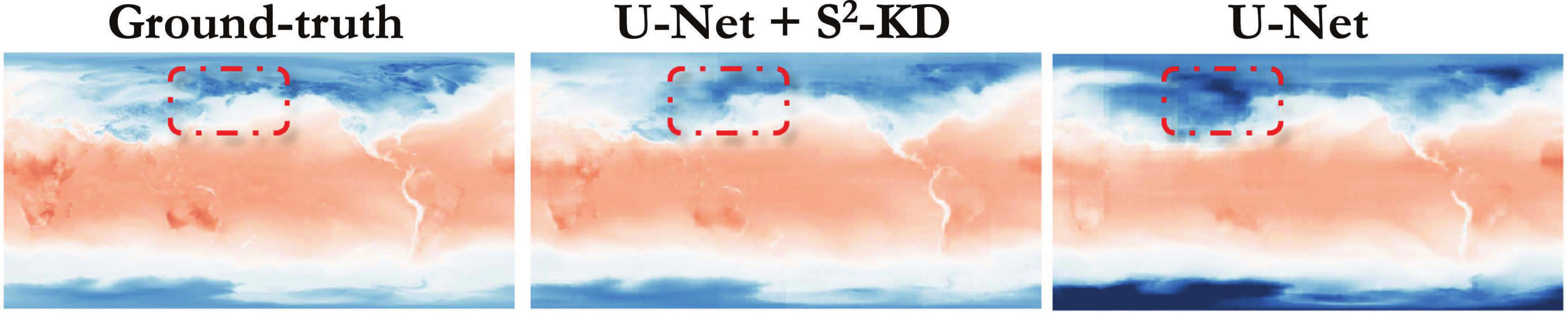} 
	\caption{Qualitative comparison of prediction results on the WeatherBench dataset.
	}
	\label{fig:qualitative_comparison}
	
\end{figure}

To provide a more intuitive understanding of the benefits of our proposed \method{} framework, we present a qualitative comparison of prediction results in Figure~\ref{fig:qualitative_comparison}. The figure displays three panels: the ground-truth weather state, the prediction from a lightweight U-Net student distilled with \method{}, and the prediction from an identical U-Net trained with a standard baseline method.

The ground-truth image (left panel) exhibits a high degree of complexity, characterized by intricate, fine-grained vortex structures and sharp gradients, particularly in the high-latitude region highlighted by the red dashed box. These features represent the high-frequency components of the atmospheric dynamics, which are notoriously difficult for compact models to capture.

The prediction from the baseline U-Net (right panel) starkly illustrates the challenge. The result is overly smooth and blurry, indicating a significant loss of high-frequency spectral information. The detailed structures within the highlighted box are smeared into an almost uniform, indistinct patch. Furthermore, the image is plagued by visible artifacts, such as unnatural blockiness and horizontal banding, which betray a superficial, pixel-level pattern imitation rather than a genuine understanding of the underlying physical processes. This outcome is a clear manifestation of the "semantic vacuum" we identified: the model learns \textit{what} the general pattern looks like (cold poles, warm equator) but remains blind to \textit{why} and \textit{how} the specific, coherent structures form.

In stark contrast, the prediction from our \method{}-enhanced student (middle panel) demonstrates a remarkable improvement. The overall clarity and sharpness are significantly closer to the ground truth. Crucially, within the highlighted region, the model successfully reconstructs the complex vortex structures with impressive fidelity. This visual evidence validates the dual-component design of our \method{} loss. The \textbf{Spectral Alignment Loss ($L_{\text{spectral}}$)} has effectively forced the student to preserve high-frequency details, preventing the blurry output seen in the baseline. More profoundly, the \textbf{Semantic Alignment Loss ($L_{\text{semantic}}$)} has endowed the student with the causal and structural knowledge distilled from the privileged, text-informed teacher. As a result, the student's prediction is not merely a collection of accurate pixels but a \textbf{semantically coherent} whole, representing a physically plausible weather state.

\section{Conclusion}
In this work, we introduced \method{}, a novel knowledge distillation framework that enriches lightweight spatiotemporal forecasting models with semantic and causal understanding, moving beyond simple pixel-level mimicry. By distilling unified semantic-spectral knowledge from a privileged, text-informed multimodal teacher into a vision-only student, \method{} significantly boosts prediction performance on diverse benchmarks, enabling simple models to approach the accuracy of massive, state-of-the-art counterparts.
\bibliography{aaai2026}

\begin{thebibliography}{45}
\providecommand{\natexlab}[1]{#1}

\bibitem[{Bi et~al.(2023)Bi, Xie, Zhang, Chen, Gu, and Tian}]{bi2023accurate}
Bi, K.; Xie, L.; Zhang, H.; Chen, X.; Gu, X.; and Tian, Q. 2023.
\newblock Accurate medium-range global weather forecasting with 3D neural
  networks.
\newblock \emph{Nature}, 619(7970): 533--538.

\bibitem[{Blikstein(2013)}]{blikstein2013multimodal}
Blikstein, P. 2013.
\newblock Multimodal learning analytics.
\newblock In \emph{Proceedings of the third international conference on
  learning analytics and knowledge}, 102--106.

\bibitem[{Bruna et~al.(2013)Bruna, Zaremba, Szlam, and
  LeCun}]{bruna2013spectral}
Bruna, J.; Zaremba, W.; Szlam, A.; and LeCun, Y. 2013.
\newblock Spectral networks and locally connected networks on graphs.
\newblock \emph{arXiv preprint arXiv:1312.6203}.

\bibitem[{Chen et~al.(2023{\natexlab{a}})Chen, Zhong, Zhang, Cheng, Xu, Qi, and
  Li}]{chen2023fuxi}
Chen, L.; Zhong, X.; Zhang, F.; Cheng, Y.; Xu, Y.; Qi, Y.; and Li, H.
  2023{\natexlab{a}}.
\newblock FuXi: A cascade machine learning forecasting system for 15-day global
  weather forecast.
\newblock \emph{npj climate and atmospheric science}, 6(1): 190.

\bibitem[{Chen et~al.(2021)Chen, Liu, Zhao, and Jia}]{chen2021distilling}
Chen, P.; Liu, S.; Zhao, H.; and Jia, J. 2021.
\newblock Distilling knowledge via knowledge review.
\newblock In \emph{Proceedings of the IEEE/CVF conference on computer vision
  and pattern recognition}, 5008--5017.

\bibitem[{Chen et~al.(2022)Chen, Li, Yang, Xie, Liu, Ma, Liu, and Tian}]{cicc}
Chen, Y.; Li, Z.; Yang, Y.; Xie, L.; Liu, Y.; Ma, L.; Liu, S.; and Tian, G.
  2022.
\newblock CICC: Channel Pruning via the Concentration of Information and
  Contributions of Channels.
\newblock In \emph{BMVC}, 243.

\bibitem[{Chen et~al.(2023{\natexlab{b}})Chen, Ren, Wang, Fang, Sun, and
  Li}]{chen2023contiformer}
Chen, Y.; Ren, K.; Wang, Y.; Fang, Y.; Sun, W.; and Li, D. 2023{\natexlab{b}}.
\newblock ContiFormer: Continuous-time transformer for irregular time series
  modeling.
\newblock In \emph{NeurIPS}.

\bibitem[{Chen and Wang(2024)}]{chen2024effective}
Chen, Y.; and Wang, Z. 2024.
\newblock An effective information theoretic framework for channel pruning.
\newblock \emph{arXiv preprint arXiv:2408.16772}.

\bibitem[{Fahlman and Fern{\'a}ndez(2022)}]{fahlman2022long}
Fahlman, S.; and Fern{\'a}ndez, R. 2022.
\newblock Long-term 3D MHD simulations of black hole accretion discs formed in
  neutron star mergers.
\newblock \emph{Monthly Notices of the Royal Astronomical Society}, 513(2):
  2689--2707.

\bibitem[{Fan et~al.(2020)Fan, Jiang, Zhang, Wang, and Lai}]{fan2020long}
Fan, H.; Jiang, J.; Zhang, C.; Wang, X.; and Lai, Y.-C. 2020.
\newblock Long-term prediction of chaotic systems with machine learning.
\newblock \emph{Physical Review Research}, 2(1): 012080.

\bibitem[{Gao et~al.(2025)Gao, Wu, Shu, Dong, Xu, Chen, Yan, Wen, Hu, Wang
  et~al.}]{gao2025oneforecast}
Gao, Y.; Wu, H.; Shu, R.; Dong, H.; Xu, F.; Chen, R.; Yan, Y.; Wen, Q.; Hu, X.;
  Wang, K.; et~al. 2025.
\newblock OneForecast: A Universal Framework for Global and Regional Weather
  Forecasting.
\newblock \emph{arXiv preprint arXiv:2502.00338}.

\bibitem[{Gao et~al.(2022)Gao, Tan, Wu, and Li}]{gao2022simvp}
Gao, Z.; Tan, C.; Wu, L.; and Li, S.~Z. 2022.
\newblock Simvp: Simpler yet better video prediction.
\newblock In \emph{Proceedings of the IEEE/CVF conference on computer vision
  and pattern recognition}, 3170--3180.

\bibitem[{Gou et~al.(2021)Gou, Yu, Maybank, and Tao}]{gou2021knowledge}
Gou, J.; Yu, B.; Maybank, S.~J.; and Tao, D. 2021.
\newblock Knowledge distillation: A survey.
\newblock \emph{International Journal of Computer Vision}, 129(6): 1789--1819.

\bibitem[{Guibas et~al.(2021)Guibas, Mardani, Li, Tao, Anandkumar, and
  Catanzaro}]{guibas2021adaptive}
Guibas, J.; Mardani, M.; Li, Z.; Tao, A.; Anandkumar, A.; and Catanzaro, B.
  2021.
\newblock Adaptive fourier neural operators: Efficient token mixers for
  transformers.
\newblock \emph{arXiv preprint arXiv:2111.13587}.

\bibitem[{He et~al.(2016)He, Zhang, Ren, and Sun}]{he2016deep}
He, K.; Zhang, X.; Ren, S.; and Sun, J. 2016.
\newblock Deep residual learning for image recognition.
\newblock In \emph{Proceedings of the IEEE conference on computer vision and
  pattern recognition}, 770--778.

\bibitem[{Hinton, Vinyals, and Dean(2015)}]{hinton2015distilling}
Hinton, G.; Vinyals, O.; and Dean, J. 2015.
\newblock Distilling the knowledge in a neural network.
\newblock \emph{arXiv preprint arXiv:1503.02531}.

\bibitem[{Kingma and Ba(2014)}]{kingma2014adam}
Kingma, D.~P.; and Ba, J. 2014.
\newblock Adam: A method for stochastic optimization.
\newblock \emph{arXiv preprint arXiv:1412.6980}.

\bibitem[{Kurth et~al.(2023)Kurth, Subramanian, Harrington, Pathak, Mardani,
  Hall, Miele, Kashinath, and Anandkumar}]{kurth2023fourcastnet}
Kurth, T.; Subramanian, S.; Harrington, P.; Pathak, J.; Mardani, M.; Hall, D.;
  Miele, A.; Kashinath, K.; and Anandkumar, A. 2023.
\newblock Fourcastnet: Accelerating global high-resolution weather forecasting
  using adaptive fourier neural operators.
\newblock In \emph{Proceedings of the platform for advanced scientific
  computing conference}, 1--11.

\bibitem[{Li et~al.(2020)Li, Kovachki, Azizzadenesheli, Liu, Bhattacharya,
  Stuart, and Anandkumar}]{li2020fourier}
Li, Z.; Kovachki, N.; Azizzadenesheli, K.; Liu, B.; Bhattacharya, K.; Stuart,
  A.; and Anandkumar, A. 2020.
\newblock Fourier neural operator for parametric partial differential
  equations.
\newblock \emph{arXiv preprint arXiv:2010.08895}.

\bibitem[{Liu et~al.(2023)Liu, Li, Wu, and Lee}]{liu2023visual}
Liu, H.; Li, C.; Wu, Q.; and Lee, Y.~J. 2023.
\newblock Visual instruction tuning.
\newblock \emph{Advances in neural information processing systems}, 36:
  34892--34916.

\bibitem[{Lu et~al.(2024)Lu, Liu, Zhang, Wang, Dong, Liu, Sun, Ren, Li, Yang
  et~al.}]{lu2024deepseek}
Lu, H.; Liu, W.; Zhang, B.; Wang, B.; Dong, K.; Liu, B.; Sun, J.; Ren, T.; Li,
  Z.; Yang, H.; et~al. 2024.
\newblock Deepseek-vl: towards real-world vision-language understanding.
\newblock \emph{arXiv preprint arXiv:2403.05525}.

\bibitem[{Mirzadeh et~al.(2020)Mirzadeh, Farajtabar, Li, Levine, Matsukawa, and
  Ghasemzadeh}]{mirzadeh2020improved}
Mirzadeh, S.~I.; Farajtabar, M.; Li, A.; Levine, N.; Matsukawa, A.; and
  Ghasemzadeh, H. 2020.
\newblock Improved knowledge distillation via teacher assistant.
\newblock In \emph{Proceedings of the AAAI conference on artificial
  intelligence}, volume~34, 5191--5198.

\bibitem[{Murata et~al.(2023)Murata, Okubo, Minematsu, Taniguchi, and
  Shimada}]{murata2023recurrent}
Murata, R.; Okubo, F.; Minematsu, T.; Taniguchi, Y.; and Shimada, A. 2023.
\newblock Recurrent neural network-fitnets: improving early prediction of
  student performanceby time-series knowledge distillation.
\newblock \emph{Journal of Educational Computing Research}, 61(3): 639--670.

\bibitem[{Park et~al.(2019)Park, Kim, Lu, and Cho}]{park2019relational}
Park, W.; Kim, D.; Lu, Y.; and Cho, M. 2019.
\newblock Relational knowledge distillation.
\newblock In \emph{Proceedings of the IEEE/CVF conference on computer vision
  and pattern recognition}, 3967--3976.

\bibitem[{Pechyony and Vapnik(2010)}]{pechyony2010theory}
Pechyony, D.; and Vapnik, V. 2010.
\newblock On the theory of learnining with privileged information.
\newblock \emph{Advances in neural information processing systems}, 23.

\bibitem[{Phuong and Lampert(2019)}]{phuong2019towards}
Phuong, M.; and Lampert, C. 2019.
\newblock Towards understanding knowledge distillation.
\newblock In \emph{International conference on machine learning}, 5142--5151.
  PMLR.

\bibitem[{Radford et~al.(2021)Radford, Kim, Hallacy, Ramesh, Goh, Agarwal,
  Sastry, Askell, Mishkin, Clark et~al.}]{radford2021learning}
Radford, A.; Kim, J.~W.; Hallacy, C.; Ramesh, A.; Goh, G.; Agarwal, S.; Sastry,
  G.; Askell, A.; Mishkin, P.; Clark, J.; et~al. 2021.
\newblock Learning transferable visual models from natural language
  supervision.
\newblock In \emph{International conference on machine learning}, 8748--8763.
  PmLR.

\bibitem[{Ramachandram and Taylor(2017)}]{ramachandram2017deep}
Ramachandram, D.; and Taylor, G.~W. 2017.
\newblock Deep multimodal learning: A survey on recent advances and trends.
\newblock \emph{IEEE signal processing magazine}, 34(6): 96--108.

\bibitem[{Rasp et~al.(2020)Rasp, Dueben, Scher, Weyn, Mouatadid, and
  Thuerey}]{rasp2020weatherbench}
Rasp, S.; Dueben, P.~D.; Scher, S.; Weyn, J.~A.; Mouatadid, S.; and Thuerey, N.
  2020.
\newblock WeatherBench: a benchmark data set for data-driven weather
  forecasting.
\newblock \emph{Journal of Advances in Modeling Earth Systems}, 12(11):
  e2020MS002203.

\bibitem[{Ronneberger, Fischer, and Brox(2015)}]{ronneberger2015u}
Ronneberger, O.; Fischer, P.; and Brox, T. 2015.
\newblock U-net: Convolutional networks for biomedical image segmentation.
\newblock In \emph{Medical image computing and computer-assisted
  intervention--MICCAI 2015: 18th international conference, Munich, Germany,
  October 5-9, 2015, proceedings, part III 18}, 234--241. Springer.

\bibitem[{Shi et~al.(2015)Shi, Chen, Wang, Yeung, Wong, and
  Woo}]{shi2015convolutional}
Shi, X.; Chen, Z.; Wang, H.; Yeung, D.-Y.; Wong, W.-K.; and Woo, W.-c. 2015.
\newblock Convolutional LSTM network: A machine learning approach for
  precipitation nowcasting.
\newblock \emph{Advances in neural information processing systems}, 28.

\bibitem[{Sorjamaa et~al.(2007)Sorjamaa, Hao, Reyhani, Ji, and
  Lendasse}]{Sorjamaa2007MethodologyFL}
Sorjamaa, A.; Hao, J.; Reyhani, N.; Ji, Y.; and Lendasse, A. 2007.
\newblock Methodology for long-term prediction of time series.
\newblock \emph{Neurocomputing}, 70(16-18): 2861--2869.

\bibitem[{Stanton et~al.(2021)Stanton, Izmailov, Kirichenko, Alemi, and
  Wilson}]{stanton2021does}
Stanton, S.; Izmailov, P.; Kirichenko, P.; Alemi, A.~A.; and Wilson, A.~G.
  2021.
\newblock Does knowledge distillation really work?
\newblock \emph{Advances in neural information processing systems}, 34:
  6906--6919.

\bibitem[{Tolstikhin et~al.(2021)Tolstikhin, Houlsby, Kolesnikov, Beyer, Zhai,
  Unterthiner, Yung, Steiner, Keysers, Uszkoreit et~al.}]{tolstikhin2021mlp}
Tolstikhin, I.~O.; Houlsby, N.; Kolesnikov, A.; Beyer, L.; Zhai, X.;
  Unterthiner, T.; Yung, J.; Steiner, A.; Keysers, D.; Uszkoreit, J.; et~al.
  2021.
\newblock Mlp-mixer: An all-mlp architecture for vision.
\newblock \emph{Advances in neural information processing systems}, 34:
  24261--24272.

\bibitem[{Wang et~al.(2020)Wang, Ma, Wang, Jin, Wang, Tang, Jia, and
  Yu}]{wang2020traffic}
Wang, X.; Ma, Y.; Wang, Y.; Jin, W.; Wang, X.; Tang, J.; Jia, C.; and Yu, J.
  2020.
\newblock Traffic flow prediction via spatial temporal graph neural network.
\newblock In \emph{TheWebConf}, 1082--1092.

\bibitem[{Wang, Han, and Chen(2025)}]{tcfi}
Wang, Y.; Han, R.; and Chen, Y. 2025.
\newblock TCFI: Topology-Consistent Pruning with Fisher Information for
  Efficient Medical Image Segmentation.
\newblock In \emph{2025 IEEE International Conference on Multimedia and Expo
  (ICME)}, 1--6.

\bibitem[{Wang et~al.(2022)Wang, Wu, Zhang, Gao, Wang, Philip, and
  Long}]{wang2022predrnn}
Wang, Y.; Wu, H.; Zhang, J.; Gao, Z.; Wang, J.; Philip, S.~Y.; and Long, M.
  2022.
\newblock Predrnn: A recurrent neural network for spatiotemporal predictive
  learning.
\newblock \emph{IEEE Transactions on Pattern Analysis and Machine
  Intelligence}, 45(2): 2208--2225.

\bibitem[{Wu et~al.(2025{\natexlab{a}})Wu, Gao, Shu, Han, Xu, Zhu, Wen, Wu,
  Wang, and Huang}]{wu2025turb}
Wu, H.; Gao, Y.; Shu, R.; Han, Z.; Xu, F.; Zhu, Z.; Wen, Q.; Wu, X.; Wang, K.;
  and Huang, X. 2025{\natexlab{a}}.
\newblock Turb-L1: Achieving Long-term Turbulence Tracing By Tackling Spectral
  Bias.
\newblock \emph{arXiv preprint arXiv:2505.19038}.

\bibitem[{Wu et~al.(2025{\natexlab{b}})Wu, Gao, Shu, Wang, Gou, Wu, Liu, He,
  Cao, Fang, Shi, Tao, Song, Ji, Xiang, Sun, Li, Xu, Dong, Wang, Zhang, Zhao,
  Wu, Wen, Chen, and Huang}]{wu2025triton}
Wu, H.; Gao, Y.; Shu, R.; Wang, K.; Gou, R.; Wu, C.; Liu, X.; He, J.; Cao, S.;
  Fang, J.; Shi, X.; Tao, F.; Song, Q.; Ji, S.; Xiang, Y.; Sun, Y.; Li, J.; Xu,
  F.; Dong, H.; Wang, H.; Zhang, F.; Zhao, P.; Wu, X.; Wen, Q.; Chen, D.; and
  Huang, X. 2025{\natexlab{b}}.
\newblock Advanced long-term earth system forecasting by learning the
  small-scale nature.
\newblock \emph{arXiv preprint arXiv:2505.19432}.

\bibitem[{Wu et~al.(2024{\natexlab{a}})Wu, Liang, Xiong, Zhou, Huang, Wang, and
  Wang}]{wu2024earthfarsser}
Wu, H.; Liang, Y.; Xiong, W.; Zhou, Z.; Huang, W.; Wang, S.; and Wang, K.
  2024{\natexlab{a}}.
\newblock Earthfarsser: Versatile Spatio-Temporal Dynamical Systems Modeling in
  One Model.
\newblock In \emph{Proceedings of the AAAI Conference on Artificial
  Intelligence}, volume~38, 15906--15914.

\bibitem[{Wu et~al.(2024{\natexlab{b}})Wu, Wang, Wang, Wang, Tao, Chen, Hua,
  Luo et~al.}]{wu2024prometheus}
Wu, H.; Wang, H.; Wang, K.; Wang, W.; Tao, Y.; Chen, C.; Hua, X.-S.; Luo, X.;
  et~al. 2024{\natexlab{b}}.
\newblock Prometheus: Out-of-distribution fluid dynamics modeling with
  disentangled graph ode.
\newblock In \emph{Forty-first International Conference on Machine Learning}.

\bibitem[{Wu et~al.(2023)Wu, Wang, Liang, Zhou, Huang, Xiong, and
  Wang}]{wu2023earthfarseer}
Wu, H.; Wang, S.; Liang, Y.; Zhou, Z.; Huang, W.; Xiong, W.; and Wang, K. 2023.
\newblock Earthfarseer: Versatile Spatio-Temporal Dynamical Systems Modeling in
  One Model.
\newblock \emph{AAAI2024}.

\bibitem[{Wu et~al.(2024{\natexlab{c}})Wu, Wen, Zhang, Xia, Liang, Zheng, Wen,
  and Wang}]{wu2024dynst}
Wu, H.; Wen, H.; Zhang, G.; Xia, Y.; Liang, Y.; Zheng, Y.; Wen, Q.; and Wang,
  K. 2024{\natexlab{c}}.
\newblock Dynst: Dynamic sparse training for resource-constrained
  spatio-temporal forecasting.
\newblock \emph{arXiv preprint arXiv:2403.02914}.

\bibitem[{Wu et~al.(2024{\natexlab{d}})Wu, Xu, Chen, Hua, Luo, and
  Wang}]{wu2024pastnet}
Wu, H.; Xu, F.; Chen, C.; Hua, X.-S.; Luo, X.; and Wang, H. 2024{\natexlab{d}}.
\newblock Pastnet: Introducing physical inductive biases for spatio-temporal
  video prediction.
\newblock In \emph{Proceedings of the 32nd ACM International Conference on
  Multimedia}, 2917--2926.

\bibitem[{Zhang et~al.(2021)Zhang, Liu, Sun, and Shah}]{zhang2021graph}
Zhang, S.; Liu, Y.; Sun, Y.; and Shah, N. 2021.
\newblock Graph-less neural networks: Teaching old mlps new tricks via
  distillation.
\newblock \emph{arXiv preprint arXiv:2110.08727}.

\end{thebibliography}

\end{document}